# Splenomegaly Segmentation on Multi-modal MRI using Deep Convolutional Networks

Yuankai Huo*, Zhoubing Xu, Shunxing Bao, Camilo Bermudez, Hyeonsoo Moon, Prasanna Parvathaneni, Tamara K. Moyo, Michael R. Savona, Albert Assad, Richard G. Abramson, and Bennett A. Landman

*Abstract*— The findings of splenomegaly, abnormal enlargement of the spleen, is a non-invasive clinical biomarker for liver and spleen disease. Automated segmentation methods are essential to efficiently quantify splenomegaly from clinically acquired abdominal magnetic resonance imaging (MRI) scans. However, the task is challenging due to (1) large anatomical and spatial variations of splenomegaly, (2) large inter- and intra-scan intensity variations on multi-modal MRI, and (3) limited numbers of labeled splenomegaly scans. In this paper, we propose the Splenomegaly Segmentation Network (SS-Net) to introduce the deep convolutional neural network (DCNN) approaches in multi-modal MRI splenomegaly segmentation. Large convolutional kernel layers were used to address the spatial and anatomical variations, while the conditional generative adversarial networks (GAN) were employed to leverage the segmentation performance of SS-Net in an end-to-end manner. A clinically acquired cohort containing both T1-weighted (T1w) and T2-weighted (T2w) MRI splenomegaly scans was used to train and evaluate the performance of multi-atlas segmentation (MAS), 2D DCNN networks, and a 3D DCNN network. From the experimental results, the DCNN methods achieved superior performance to the state-of-the-art MAS method. The proposed SS-Net method achieved the highest median and mean Dice scores among investigated baseline DCNN methods.

*Index Terms*— Spleen Segmentation, MRI, Deep Convolutional Neural Network, multi-contrast, splenomegaly

## I. INTRODUCTION

SPLENOMEGALY, abnormal enlargement of the spleen, is associated with abnormal red blood cell destruction, which has been used as a clinical biomarker associated with liver disease [1], infection [2] and cancer [3]. Non-invasive medical imaging techniques (e.g., ultrasound [4, 5], computed tomography (CT) [6-9], and magnetic resonance imaging (MRI) [10, 11]) have been used for clinical and scientific investigations on splenomegaly. Manually tracing a spleen on two-dimensional (2D) slices from a three-dimensional (3D) volume has been regarded as the gold standard of imaging-based spleen volume estimation [11]. However, manual tracing is resource and time consuming and not routinely feasible for large cohorts. To alleviate the manual efforts in clinical practice, many previous endeavors have been conducted. One direction is to perform quick one-dimensional (1D) measurements (e.g., splenic length, width, thickness) [7] to estimate 3D volume size using regression model. Another direction has sought to develop fully automated spleen segmentation methods [12] including, but not limited to, intensity-based methods [13], shape/contour-based models [14], graph cuts[15], learning based models [16], and atlas-based approaches [17].

Historically, methods were mostly developed for CT spleen segmentation since CT is the *de facto* standard abdominal imaging modality [18]. One key benefit of using CT for spleen segmentation is that the intensities are scaled by tissue-specified Hounsfield Unit (HU) [19]. However, it hindered the generalization to intensity-based [20] and feature-based segmentation methods [21] on non-scaled imaging modalities (e.g., MRI). In the past decades, MRI has been widely used in clinical scenarios as a radiation risk free imaging modality and for improved soft tissue contrast [22]. Therefore, MRI based spleen volume estimation is appealing in clinical application. Yet, relatively few previous spleen segmentation methods have been proposed for MRI. A dual-space clustering technique has been proposed by Farraher et al [23]. Wu et al. combined Gabor

This research was supported by NSF CAREER 1452485 (Landman), NIH grants 5R21EY024036 (Landman), R01EB017230 (Landman), 1R21NS064534 (Prince/Landman), 1R01NS070906 (Pham), 2R01EB006136 (Dawant), 1R03EB012461 (Landman), P30 CA068485-VICC NCI Cancer Center Support Grant (Savona, Abramson), and R01NS095291 (Dawant). This research was also supported by the Vanderbilt-Incyte Research Alliance Grant (Savona/ Abramson/ Landman). This research was conducted with the support from Intramural Research Program, National Institute on Aging, NIH. This study was also supported by NIH 5R01NS056307, 5R21NS082891 and in part using the resources of the Advanced Computing Center for Research and Education (ACCRE) at Vanderbilt University, Nashville, TN. This project was supported in part by ViSE/VICTR VR3029 and the National Center for Research Resources, Grant UL1 RR024975-01, and is now at the National Center for Advancing Translational Sciences, Grant 2 UL1 TR000445-06. The content is solely the responsibility of the authors and does not necessarily represent the official views of the NIH. We appreciate the NIH S10 Shared Instrumentation Grant 1S10OD020154-01 (Smith), Vanderbilt IDEAS grant (Holly-Bockelmann, Walker, Meliler, Palmeri, Weller), and ACCRE's Big Data TIPs grant from Vanderbilt University.

*Y. Huo is with the Department of Electrical Engineering and Computer Science, Vanderbilt University, Nashville, TN 37235 USA (e-mail: yuiankai.huo@vanderbilt.edu)

Z. Xu, S. Bao, H. Moon, P. Parvathaneni, and B. A. Landman are with the Department of Electrical Engineering and Computer Science, Vanderbilt University, TN 37235 USA

C. Bermudez is with the Department of Biomedical Engineering, Vanderbilt University, TN 37235 USA

T. K. Moyo and M. R. Savona are with the Department of Medicine, Vanderbilt University Medical Center. TN 37235 USA

A. Assad is with Incyte Corporation, Delaware 19803 USA

R. G. Abramson is with the Department of Radiology and Radiological Science, Vanderbilt University Medical Center. TN 37235 USA

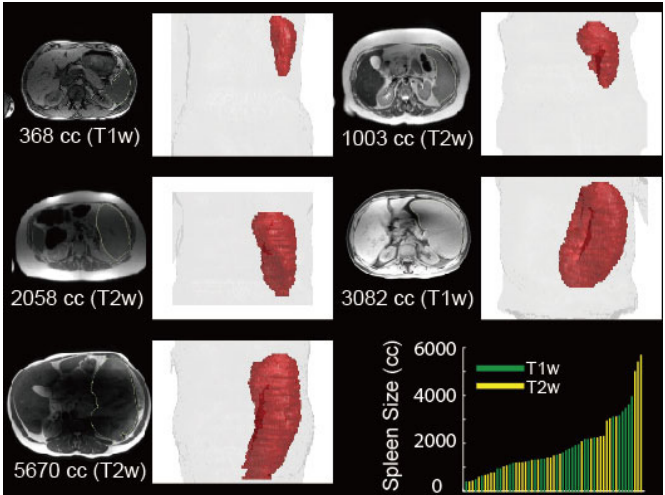

Fig. 1. This figure illustrates the large anatomical and spatial variations of multi-modal splenomegaly MRI scans. The intensity images in axial view and 3D renderings of spleen are shown with the modality and spleen volume size in cubic centimeter (cc). The lower right panel summarizes the spleen volume size and modality for all scans in this study,

features with snake method for MRI spleen segmentation [24]. Behard et al. developed a segmentation method using neural network and recursive watershed [16]. Pauly et al. proposed to use supervised regression-based segmentation for MRI Dixon sequences [25]. However, very limited previous work has been conducted on MRI splenomegaly segmentation, which not only faces the intensity variations but also deals with large spatial and anatomical variations. Recently, Huo et al. introduced multi-atlas segmentation (MAS) methods in splenomegaly segmentation on MRI [26, 27]. In that method, the L-SIMPLE atlas selection method was designed to overcome the large anatomical and spatial variations of splenomegaly. Meanwhile, the key techniques in MAS (e.g., registration and label fusion) relied on image context rather than absolute image intensities, which made MAS a reasonable solution for multi-modal MRI scenarios. As a result, L-SIMPLE MAS achieved decent overall segmentation accuracy (median Dice similarity coefficient (DSC)>0.9). However, the outliers (e.g., whose DSC<0.8) hinder accurate clinical estimation of splenomegaly.

In recent years, deep convolutional neural network (DCNN) image segmentation methods have been proposed for semantic segmentation including, but not limited to FCN [28], U-Net [29], SegNet [30], DeepLab [31], DeconvNet [32]. These methods have been successfully applied on abdominal medical image segmentation tasks (e.g., pancreas [33, 34], liver [35, 36], multi-organ [37]). However, to the best of our knowledge, few previous DCNN methods have been proposed for MRI spleen segmentation, and no previous DCNN methods have been proposed for splenomegaly segmentation on multi-modal MRI (using both T1-weighted (T1w) and T2-weighted (T2w) scans). To perform multi-modal MRI splenomegaly segmentation, three challenges need to be addressed: (1) large spatial and anatomical variations for splenomegaly (**Fig. 1**), (2) large inter-modality (T1w vs. T2w) and intra-modality (within T1w or T2w) intensity variations in multi-modal MRI, and (3) limited number of labeled training data as the splenomegaly MRI scans are more difficult to be acquired than normal spleen MRI scans. Therefore, proposed DCNN methods should be able to deal with large spatial and inter/intra-modality variations of multi-modal splenomegaly scans with limited training scans.

In this paper, we propose the splenomegaly segmentation network (SS-Net) to leverage the segmentation performance with following features: (1) large convolutional kernels were used in the skip connection layers for the large spleen, (2) adversarial networks were employed as a discriminator to leverage the segmentation performance for an end-to-end training, and (3) 2D+ multi-view training was used to improve the splenomegaly segmentation. We evaluate the performance of the proposed SS-Net as well as prevalent DCNN segmentation frameworks on clinically acquired multi-modal (both T1w and T2w) splenomegaly cohorts.

The work extends our previous conference paper [38] with the following new efforts: (1) the methodology on network structures, parameters and loss functions are presented in greater details, (2) new 2D and 3D networks and details of multi-view fusion are provided for evaluation, and (3) more comprehensive analyses (e.g., 4-fold cross validation on entire cohort, sensitivity analyses on adversarial loss and hyper-parameters) were presented. For the methodology, our previous conference paper used dummy large 2D kernels (consisting of two 1D kernels) for skip-connection layers (same as [39]), while the real large 2D kernels were used in this work. Meanwhile, more training epochs had been performed in this work (50 epochs) compared with our previous conference version (10 epochs). The source code of the proposed SS-Net is made publicly available (https://github.com/MASILab/SSNet).

## II. RELATED WORKS

### A. Global Convolutional Network

Medical image segmentation using DCNN can be regarded as a per-voxel classification task, which assigns a class label for each voxel. As a classifier, the property of spatial invariance is important in network design. However, different from a canonical whole image (or whole volume) classification task, the spatial location of each voxel and spatial relationship between voxels are also essential in precise image segmentation (e.g., a spleen appears in the right side of the abdomen). Therefore, a proper segmentation method should balance two aspects: classification and localization. For localization, the Fully Convolutional Network (FCN) was proposed to perform image segmentation [28] and has been widely validated for different applications. The advantage of FCN is to keep spatial information across the entire network without using global pooling or dense connection.

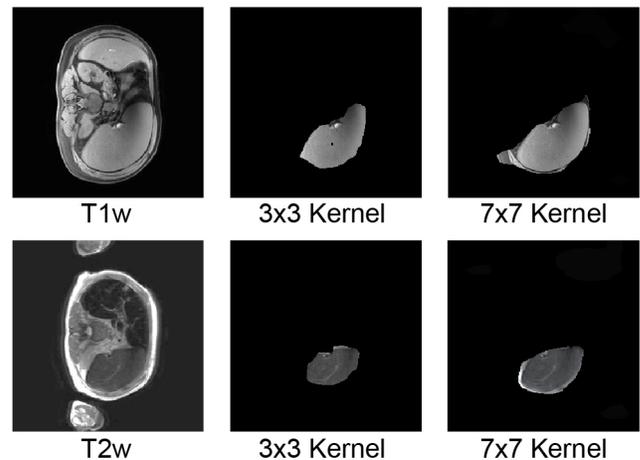

Fig. 2. This figure compares using large convolutional kernel and small convolutional kernel in splenomegaly segmentation. The upper row indicated a T1w image, while the lower row posed a T2w image. The first column was the original intensity image while the right two columns indicated masked valid field of view (FOV) in skip-connector layers. The larger kernel (7 × 7) has larger valid FOV compared with smaller kernel (3 × 3), which contained entire spleen.

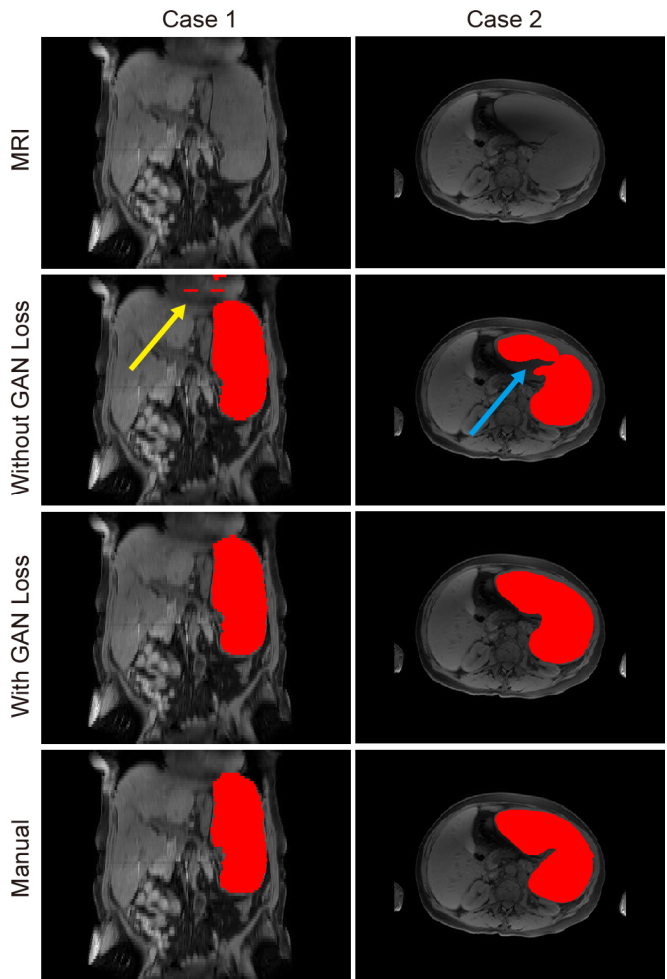

Fig. 3. This figure presents two examples of using GAN loss and without using GAN loss. The first row indicated the original MRI intensity images. The second row presented the segmentation results without using GAN loss, while the third row presented the segmentation results using GAN loss. The last row presented the manual segmentation results. The two columns indicated the two cases of using and without using GAN loss. In case 1, the spatial inconsistency was alleviated by using GAN loss. In case 2, the segmentation result was more realistic when using GAN loss.

Many previous efforts have been made of use the DCNN, especially the FCN, on abdominal organ segmentation. Zhou et al. [40] proposed fixed point model to shrink the input region using predicted segmentation masks, which led to better performance for pancreas segmentation. Meanwhile, Zhou et al. [41] introduced extra deep supervision into segmentation network, which achieved superior performance on pancreas cyst segmentation. Cai et al. [42] introduced the recurrent neural contextual learning to the pancreas segmentation by utilizing the spatial consistency among neighboring slices. These important contributions improved segmentation performance for the abdominal organ with large spatial variations (e.g., pancreas).

Recently, Peng et al. [39] revealed that the traditional small convolutional kernels (e.g., 3×3) in FCN limited the per-pixel classification accuracy. To further leverage the classification capability of FCN, the large convolutional kernels (named as global convolutional networks (GCN) [39]) in skip-connector layers were introduced to increase the valid field of view (FOV) [43]. To further improve computational efficiency, GCN used two large 1D convolutional kernels to simulate a single large 2D kernels for the skip-connector layer. The larger valid FOV in GCN provided a solution to overcome the large spatial variations for the segmenting targets. As we are facing the similar challenges for splenomegaly segmentation (large spatial and anatomical variations), we adapted the large kernel idea from GCN. However, different from GCN, full large 2D kernels were used rather than dual 1D kernels since the computational complexity was acceptable. **Fig. 2** showed a case that the larger 7×7 kernel provided larger valid field of view compared with regular 3×3 kernel.

### B. GAN loss in segmentation

Goodfellow et al. [44] proposed the generative adversarial network (GAN) to discriminate samples from empirical data and deep neural networks. The idea of the GAN was to let DCNN to perform adversarial learning, rather than only relying on the canonical loss function and regularities. Soon after, GAN was successfully applied to many computer vision tasks like image augmentation [45], super-resolution [46], style transfer [47] etc. Recently, Luc et al. [48] introduced the GAN to semantic segmentation by using it as adversarial loss functions, which not only leveraged overall segmentation performance, but also alleviated the spatial incorrectness (e.g., isolated pixels/voxels, inaccurate boundaries). The rationale was to use GAN framework to supervise the training procedure to make the segmentation results more "realistic" in an end-to-end manner, instead of performing iterative post-processing refinements like conditional random field (CRF) [49], level-set [50], active shape model (ASM) [51]. To leverage the segmentation performance and alleviate the spatial inconsistencies (**Fig. 3**) in splenomegaly segmentation, we integrated the recently proposed conditional PatchGAN [47] to our SS-Net segmentation framework.

### C. 2D and 3D Networks

There is no clear method of choosing of 2D or 3D networks for a particular segmentation task since performance depends on application scenarios, available training data, and hardware capability. Most DCNN segmentation networks were designed for 2D natural images, while many medical data in clinical practice are 3D volumes. To perform 3D segmentation, we could either perform slice-wise 2D segmentation using 2D networks or apply 3D networks for 3D segmentation directly.

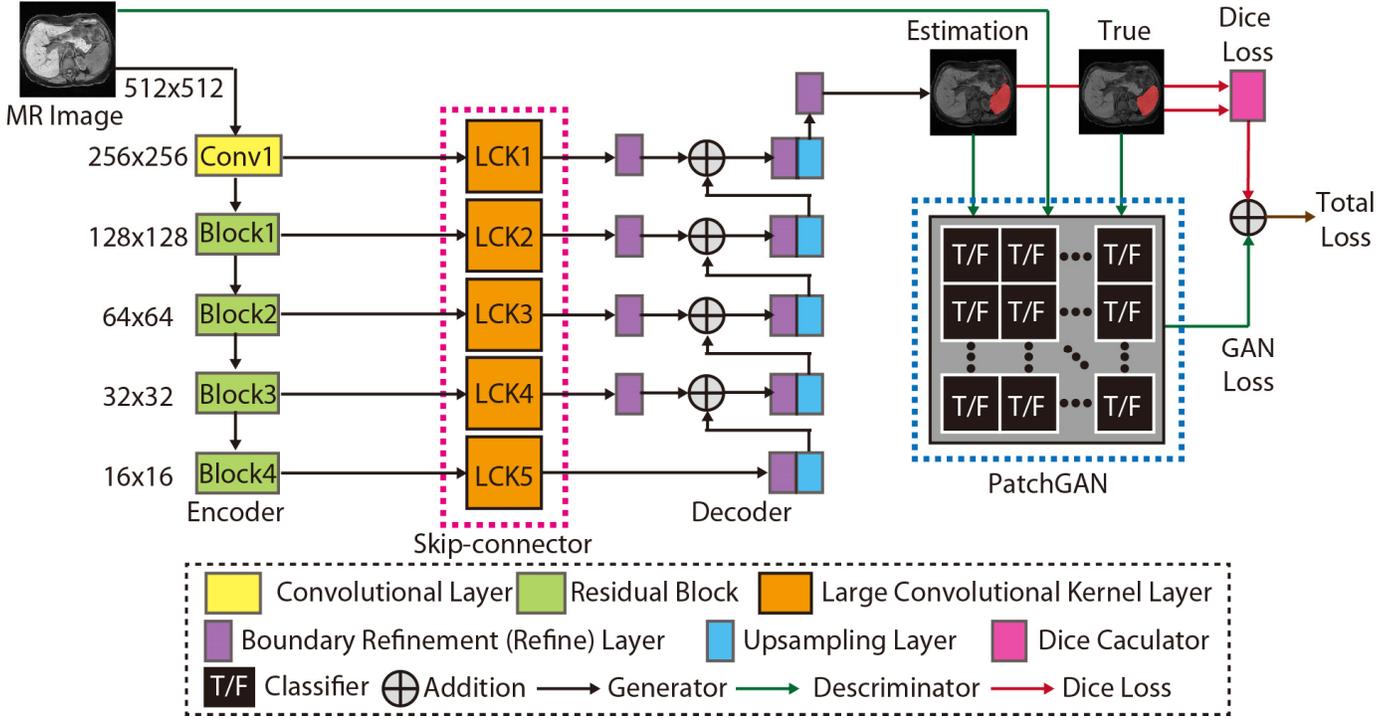

Fig. 4. This figure presented the network structure of the proposed SS-Net. From left to right, the network consisted of (1) Encoder, (2) Skip-connector, (3) Decoder, and (4) PatchGAN discriminator. The name of each network component was showed in the dashed legend box. Briefly, the ResNet50 was used as encoder. The large convolutional kernels (LCK) with kernel size 7 × 7 were used as skip-connector. The boundary refinement layers and upsampling layers were employed as decoder. Using true intensity images, true segmentation images, and fake segmentation images, the PatchGAN functioned as the discriminator to supervise the training procedure. Finally, the Dice loss and the GAN loss were combined to derive the final training loss for the proposed SS-Net.

However, the use of 3D networks is limited by GPU memory size, image resolution, and number of training volumes. Therefore, we compare the proposed method with prevalent 2D (U-Net [29], ResNet-FCN [28, 52], GCN [39]) and 3D networks (3D U-Net [53]). For simplicity, the ResNet-FCN was simplified as "ResNet" in this paper. Recently, Roth et al. [54] proposed the holistically-nested convolutional networks (HNNs), which fused the three orthogonal axial, sagittal, and coronal views to leverage the pancreas segmentation performance. The multi-view strategy achieved better segmentation performance by utilizing the information from three views. Therefore, we follow a similar perspective in the proposed method.

III. METHOD

In this paper, we propose the splenomegaly segmentation network (SS-Net), which is presented in **Fig. 4**. SS-Net is a 2D network using axial view image slices. Meanwhile, we proposed SS-Net+ method (**Fig. 5**), which used all axial, coronal, and sagittal view slices in splenomegaly segmentation.

A. Preprocessing

The intensities of every input MRI scan were normalized to 0-1 scale, whose minimal value was 0 and maximal value was 1. The highest 2.5% and lowest 2.5% intensities were excluded from the normalization to reduce the outlier effects. Then the normalized intensity image volume was resampled to 512x512x512 using bilinear interpolation, while the corresponding label volume was resampled to the same resolution using nearest neighbor interpolation. So that, we were able to obtain the same resolution (512x512) 2D slices on axial, coronal, and sagittal views. In SS-Net, we used only axial slices for each scan during training, while in SS-Net+, we used all axial, sagittal, and coronal slices.

B. SS-Net network

The network structure of SS-Net is showed in **Fig. 4**, which followed FCN framework. SS-Net consisted of four portions: (1) encoder, (2) decoder, (3) skip-connector, and (4) discriminator (**Table I**).

**Encoder:** The ResNet-50 structure [52], a ResNet framework with 50 layers, was used as encoder. The first portion ("Conv1") obtained 64 channels feature images from three input channels. The three input channels are three identical input 2D slices to simulate RGB inputs. Then four rescale blocks [52] ("Block1", "Block2", "Block3", and "Block4") were employed to obtain hieratical image features. Each rescale block contains 3 to 6 "Bottleneck" structures.

**Decoder:** The decoder of SS-Net followed the design of canonical FCN [28], in which the bilinearly upsampled ("Upsample") feature maps from previous layer were added with the feature maps from skip-connectors. The refine layers ("Refine") were implemented in the decoder before and after addition operations to further improve the segmentation

TABLE I
NETWORK STRUCTURE OF SS-NET

**Encoder:**

| | |
|---|---|
| Conv1 | Conv2d(IC=3, OC=64, KS=7, stride=2, padding=3) <br> BatchNorm2d() <br> ReLU() <br> MaxPool2d(size=3, stride=2, padding=1, dilation=1) |
| Block1 | Bottleneck+( IC=64, OC=256) <br> Bottleneck(IC=256, OC=256) <br> Bottleneck(IC=256, OC=256) |
| Block2 | Bottleneck+( IC=256, OC=512) <br> Bottleneck(IC=512, OC=512) <br> Bottleneck(IC=512, OC=512) <br> Bottleneck(IC=512, OC=512) |
| Block3 | Bottleneck+( IC=512, OC=1024) <br> Bottleneck(IC=1024, OC=1024) <br> Bottleneck(IC=1024, OC=1024) <br> Bottleneck(IC=1024, OC=1024) <br> Bottleneck(IC=1024, OC=1024) <br> Bottleneck(IC=1024, OC=1024) |
| Block4 | Bottleneck+( IC=1024,2048) <br> Bottleneck(IC=2048,2048) <br> Bottleneck(IC=2048,2048) |

**Decoder:**

| | |
|---|---|
| Refine | BatchNorm2d() <br> ReLU() <br> Conv2d(IC=2, OC=2, KS=7, stride=1, padding=1) |
| Upsample | upsample_bilinear() |

**Skip-connector:**

| | |
|---|---|
| LCK1 | Conv2d(IC=64, OC=2, KS=7, stride=1, padding=3) |
| LCK2 | Conv2d(IC=128, OC=2, KS=7, stride=1, padding=3) |
| LCK3 | Conv2d(IC=256, OC=2, KS=7, stride=1, padding=3) |
| LCK4 | Conv2d(IC=512, OC=2, KS=7, stride=1, padding=3) |
| LCK5 | Conv2d(IC=1024, OC=2, KS=7, stride=1, padding=3) |

**Discriminator:**

| | |
|---|---|
| PatchGAN | Conv2d(IC=5, OC=64, KS=4, stride=2, padding=1) <br> LeakyReLU() <br> Conv2d(IC=64, OC=128, KS=4, stride=2, padding=1) <br> BatchNorm2d() + LeakyReLU () <br> Conv2d(IC=128, OC=256, KS=4, stride=2, padding=1) <br> BatchNorm2d() + LeakyReLU () <br> Conv2d(IC=256, OC=512, KS=4, stride=1, padding=1) <br> BatchNorm2d() + LeakyReLU () <br> Conv2d(IC=512, OC=1, KS=4, stride=1, padding=1) <br> Sigmod() |

* "IC" is the input channel number, "OC" is the output channel number, "KS" is the kernel size,

performance [39]. The upsampling layers and refine layers were showed in **Table I**.

**Skip-connector:** The 2D convolutional layers with large convolutional kernels (LCK) and different number of channels were used as skip-connector with kernel size. The kernel size was empirically set to 7×7 in SS-Net according to the performance in [39] as well as the computational efficiency in practice. The number of channels for each LCK ("LCK1", "LCK2", "LCK3", "LCK4", and "LCK5") was presented in **Table I**. In the proposed SS-Net, we proposed to use real 2D large convolutional kernel (e.g., 7×7) in the skip-connector rather than using pseudo-2D large convolution kernel (consisting of two 1D large convolutional kernel, e.g., 7×1 and 1×7) in the previously proposed GCN [39]. Since the number of available MRI splenomegaly scans for training is limited, the real 2D large convolutional kernel is able to provide superior performance compared with the pseudo large kernel in GCN, and the extra computational time and memory consumption was acceptable considering the size of training cohort.

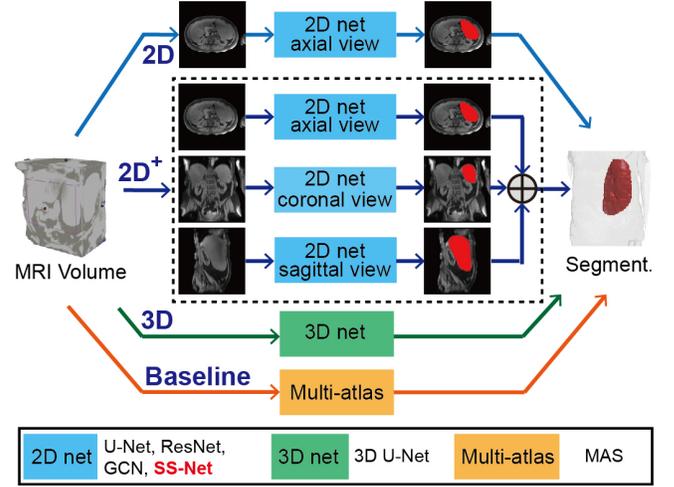

Fig. 5. This figure presented the experimental design of different methods. The 2D models were trained using only axial view images, while the 2D+ models were trained using all three views. The entire 3D volumes were used to train 3D models. The L-SIMPLE based multi-atlas segmentation (MAS) method was used as the baseline method.

**Discriminator:** We employed the PatchGAN [47] as an additional discriminator to supervise the training procedure of SS-Net. The network structure of the PatchGAN is a patch-wise classifier. Briefly, the true segmentation $Ts$ was put into the discriminator $D(\cdot)$, whose details were showed in **Table I**. Then, the input image $I$ was put into the generator $G(\cdot)$ (encoder, decoder, and skip-connector) to get the outputs for each patch $z$. After getting fake labels $a$ and true labels $b$, the Least Squares Generative Adversarial Networks (LSGAN) was employed to calculate the loss function in Eq. 1. Meanwhile, the loss function of generator was calculated in Eq. 2, where c is the value that $G(\cdot)$ for $D(\cdot)$ to incorrectly label data.

$$Loss_{GAN}(D) = \frac{1}{2}\mathbb{E}_{x \in Patch(Ts)}[(D(x) - b)^2] \\ + \frac{1}{2}\mathbb{E}_{z \in Patch(I)}\left[\left(D(G(z)) - a\right)^2\right] \quad (1)$$

$$Loss_{GAN}(G) = \frac{1}{2}\mathbb{E}_{z \in Patch(I)}[(D(G(z)) - c)^2] \quad (2)$$

### C. Loss Functions

Dice loss was used as a canonical supervision loss in this study. The traditional Dice loss was calculated slice-wise and averaged in a mini-batch. However, Dice loss might be unstable since the spleen might not appear on some slices. To alleviate the instability, we calculated the Dice loss on whole batch level rather than slice level. When calculating the Dice on whole batch level (e.g., batch size=12), we were able to improve the robustness of the Dice loss as the 2D images in a batch were randomly chosen from the entire training cohort. The formula of Dice loss is presented as $Loss_{DSC}$

$$Loss_{DSC} = -\frac{2\sum_{b=1}^{M}\sum_{i=1}^{N} T_{bi}P_{bi} + \varepsilon}{\sum_{b=1}^{M}\sum_{i=1}^{N} T_{bi}^2 + \sum_{b=1}^{M}\sum_{i=1}^{N} P_{bi}^2 + \varepsilon} \quad (3)$$

where $M$ is the number of batches while $N$ is the number of voxels in each slice. $b$ is the batch index while $i$ is the voxel index. $T$ and $P$ represents the manual segmentation (0-1 mask) and the predicted segmentation (0-1 mask). $\varepsilon$ is a small constant to avoid zero in the denominator. $\varepsilon$ was empirically set to $1e^{-7}$.

Moreover, we incorporated the GAN loss in the final loss function. We added the Dice loss with the generator loss to train the generator.

$$Loss_{ss} = Loss_{DSC} + \lambda \cdot Loss_{GAN}(G) \cdot mod(b, k) \quad (4)$$

where $\lambda$ was a constant coefficient to adjust the ratio of two components in Eq. 4. In this study, $\lambda$ was empirically set to 0.01 to balance the Dice loss and adversarial loss since the Dice loss was the major loss that we would the generator to learn from as a segmentation network. The $mod(b, k)$ is the modulus from dividing the batch index $b$ by a constant $k$. The $mod(b, k)$ is zero unless the batch index $b$ is the times of $k$. The rationale of adding this term is to alleviate the saturation issue in training discriminator, since the saturated discriminator might not provide learnable information for generator. The $k$ was empirically set to 100 in this study, which means the GAN loss was performed every 100 batches.

When training the SS-Net, $Loss_{ss}$ and $Loss_G(D)$ were used iteratively using Adam optimization [55]. Since the discriminator typically converges faster than generator, we trained the generator in each iteration, while trained the discriminator in every 100 iterations.

### D. SS-Net+ network

SS-Net only used axial view images during training since they had less anatomical and spatial variation compared with coronal and sagittal views for splenomegaly. However, other views might still provide complementary information for splenomegaly segmentation. Therefore, we proposed SS-Net+ framework, in which all axial, coronal, and sagittal views were used to train three independent SS-Nets. Then, three 3D volumetric segmentations were obtained from the three networks (**Fig. 5**). Next, we perform the union operation to fuse them to a single segmentation. To further smooth the segmentation, open and close morphological operations (radius=3) were performed to obtain the final segmentation of SS-Net+ (**Fig. 6**). Briefly, the union operations were applied to merge 3D segmentation volumes from sagittal, coronal, and axial views to a single segmentation volume. Then, the open and close morphological operations were applied to refine the final segmentation results by smoothing the segmentation boundaries and filling the holes.

## IV. DATA AND PLATFORM

60 clinically acquired whole abdominal MRI scans (32 T1w / 28 T2w) from splenomegaly 27 unique patients were used as the experimental data. Among all patients, six subjects have two T1w scans and one T2w scan, four subjects have one T1w scans and two T2w scans, two subjects have one T1w scan and one T2w scan. The remaining are the subjects only have T1w or T2w scans. The spleen size ranged from 369 cubic centimeter (cc) to 5670 cc with mean and standard deviation as 1881 cc and 219 cc. The volume size and modality of each scan was showed in the lower right pane of **Fig. 1**. Acquired scans were de-identified for the following image processing. Among the 60 scans, 45 of them (24 T1w / 21 T2w) were randomly used as training data, while the remaining 15 (8 T1w / 7 T2w) scans were used as withhold empirical validation cohort. Since the T1w and T2w scans are acquired from the same patients for the 12 subjects, we have ensured that the scans from the same patient were all used in training or testing. The dimension of MRI scans varies from $256 \times 210 \times 70$ to $512 \times 512 \times 112$ with mean resolution $441 \times ds414 \times 76$. To focus on the improvements from the networks and training strategies, we did not employ the data augmentation step (e.g., translation, rotation, flip etc.) when sampling the training and testing data.

The experiments in this study were performed using NVIDIA Titan GPU (12 GB memory) and CUDA 8.0. The training and testing were performed on a regular workstation with Intel Xeon ES-2630 V4 2.2 GHz CPU, 32G memory. The code of multi-atlas segmentation was implemented in MATLAB 2016a (www.mathworks.com), while the code of DCNN methods was implemented in Python 2.7 (www.python.org). The anaconda was (https://www.anaconda.com) used to host virtual Python environment and the corresponding library configurations. For DCNN methods, the PyTorch 0.2 version (www.pytorch.org) was used to establish the network structures and perform training.

## V. EXPERIMENTAL DESIGN

### A. Multi-atlas Segmentation

L-SIMPLE multi-atlas segmentation (MAS) pipeline [27] was employed as the baseline method to present the state-of-the-art performance for multi-modal MRI splenomegaly segmentation. Briefly, each atlas was registered to the target image using DEnsE Displacement Sampling (DEEDS) [56]. Next, an atlas selection step was performed using the expectation-maximization (EM) based L-SIMPLE method. Then, 10 selected atlases were fused to one segmentation using joint label fusion (JLF) [57]. Finally, the segmentation was refined by performing post-processing using graph cuts method [58]. Note that for the MAS method, the leave-one-subject-out strategy was used to be consistent with [27].

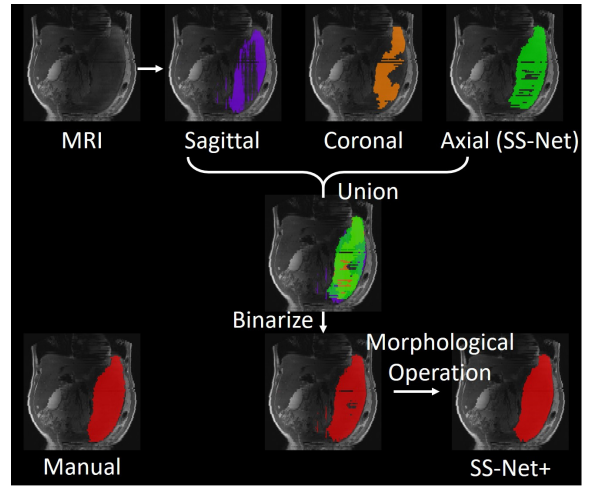

Fig. 6. This figure presented the experimental design of the multi-view fusion stage ("+" sign in **Fig. 5**). Three whole spleen segmentations were obtained from input MRI volume. Then, the union operation was used to combine three segmentation to a single whole spleen segmentation. Last, the binary 3D morphological operations (open and close) were performed to refine the final segmentation result.

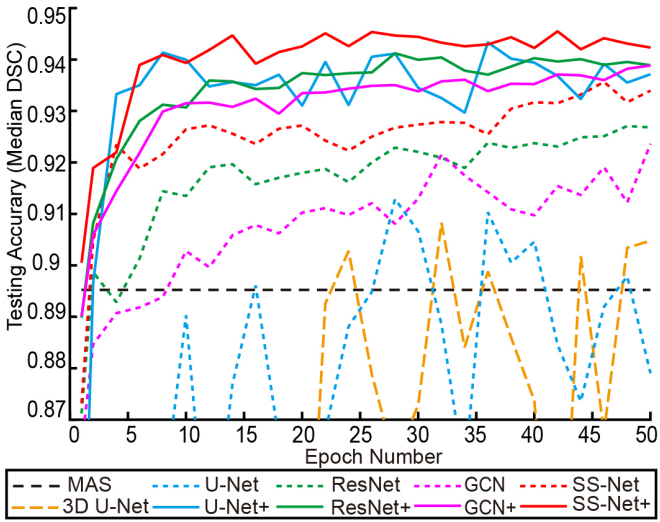

Fig. 7. The median DSC was used to evaluate the testing accuracy for different methods. The black line presented the median segmentation performance of MAS as the baseline method. The 3D and 2D methods were showed as dashed lines, while the 2D+ methods were reflected as solid lines.

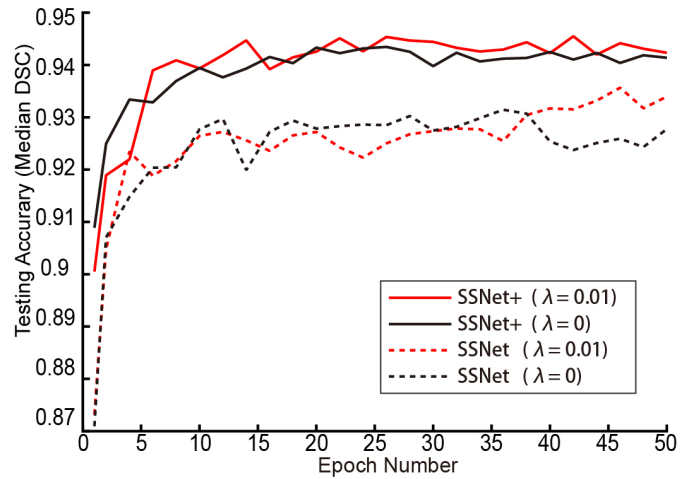

Fig. 8. The median DSC for SS-Net and SS-Net+ was provided. This figure showed the effect of using and without using GAN for SS-Net and SS-Net+ training. λ=0 indicated the GAN was not used during training, while λ=0.01 indicated the GAN was used during training. For single-view SS-Net, training with GAN achieved superior performance after epoch 38. For multi-view SS-Net+, training with GAN achieved slightly better overall performance since the multi-view integration reduced the effect of without using GAN (**Fig. 3**).

*B. 3D Model*

The widely validated 3D U-Net was used as another baseline method, whose network structure followed [53]. Briefly, each training and testing MRI volume was resampled to $176 \times 176 \times 64$, which was able to be fitted into 12GB GPU memory. The trilinear interpolation was used for resampling the intensity volume, while the nearest neighbor was used for resampling the label volume. Lastly, the output segmentation form 3D U-Net was resampled to the original scan resolution using nearest-neighbor sampling.

*C. 2D Models*

Compared from 3D model, the 2D model enabled larger resolution for each input image. Therefore, we performed data augmentation by resampling each MRI scan to 512x512x512 resolution. Then, 512 axial slices were obtained for each scan to train a 2D network.

For 2D models, we used the same training parameters for different methods: learning rate=0.00001, batch size=12, optimizer=Adam, loss function = Dice loss. The batch size of U-Net was set to 4 since that was the maximal number can be fitted into 12GB GPU memory. The learning rate was kept and same and small the Adam optimizer to ensure the stability of training across different splenomegaly segmentation deep networks. The value was also kept the same as the previous work [38] to enable us to compare new results with the previous results.

Since the training and testing were performed in 2D manner, the final 3D segmentations were obtained by stacking the 2D segmentations into 3D space. Then, the 512x512x512 outputs were resampled the to the original scan resolution using nearest neighbor interpolation.

*D. 2D+ Models*

The SS-Net+ method was introduced in the "Method" section. Using the same idea, we introduce U-Net+, ResNet+, and GCN+. We called all these methods as 2D+ models. As **Fig. 5**, The major difference between 2D+ and 2D models was that all three views (axial, coronal, sagittal) were used in the training and testing procedures. Briefly, three independent networks were trained for each method for the three views. Since each MRI scan was resampled to 512x512x512 resolution in the preprocessing, 512 axial slices, 512 coronal slices, and 512 sagittal slices were obtained for each scan to train three networks. Each individual network in the 2D+ model was identical to the 2D model. From 2D+ model, three 512x512x512 segmentation outputs were merged into a single 3D segmentation as SS-Net+. Finally, the 512x512x512 outputs were resampled the to the original scan resolution using nearest neighbor interpolation.

*E. Measurements*

The Dice similarity coefficient (DSC) was used as the major measurement to evaluate the performance of different segmentation methods. The mean surface distance (MSD) and Hausdorff distance (HD) were employed as complementary measurements to reflect the segmentation accuracy on surface manner. The differences between methods were evaluated by Wilcoxon signed rank test [59] and the difference was significant means $p<0.05$ in this paper.

## VI. RESULTS

The median testing DSC along with 50 training epochs was presented in **Fig. 7**. The results showed that the 2D and 2D+ based methods achieved higher DSC than 3D U-Net. The 2D+ based methods had higher DSC than 2D based methods. Within 2D and 2D+ family, the proposed SS-Net and SS-Net+ methods generally achieved consistent higher median DSC than U-Net, ResNet, GCN, U-Net+, ResNet+, and GCN+ along the epoch number. By fusing the multi-view results from three models (axial, coronal, and sagittal in **Fig. 6**), the SS-Net+ method achieved superior performance compared with the SS-Net. Since the network structure of SS-Net and SS-Net+ were identical, the improvements on segmentation performance came from the multi-view integration. From **Fig.9**, **Fig.10** and **Table II**, the U-Net+, ResNet+, GCN+, and SS-Net+ achieved more similar performance compared with U-Net, ResNet, GCN, and SS-Net. It demonstrated that the multi-view integration is a robust boosting strategy, which leveraged the performance of different basic methods in a consistent way.

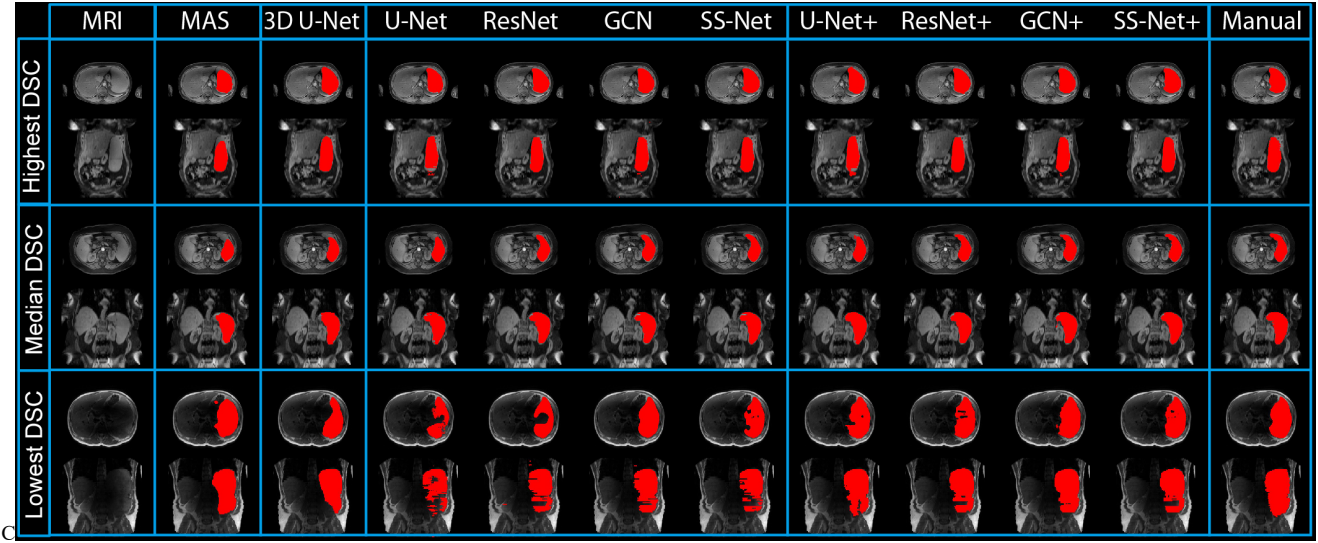

Fig. 9. The qualitative results were presented in this figure. The results from three subjects (highest, median, and lowest Dice similarity coefficient (DSC) in SS-Net+) were showed in the upper, middle, and lower rows. Different segmentation methods were showed as different columns. The multi-atlas segmentation (MAS) results were presented as baseline. Then, 3D, 2D, and 2D+ models were presented to show the performance of DCNN methods. The right-most column reflected the manual segmentation results.

The quantitative performance of using and without using PatchGAN is provided in the **Fig. 8**. The median DSC segmentation results of using PatchGAN ($\lambda=0.01$) and without using PatchGAN ($\lambda=0$) across 50 training epochs have been presented. For the single-view strategy SS-Net, the splenomegaly segmentation with GAN achieved superior performance when the number of epochs became larger. For the multi-view strategy SS-Net+, the splenomegaly segmentation with GAN had slightly better overall performance. The improvement was smaller for SS-Net+ is because that the multi-view fusion stage alleviated the segmentation performance (as shown in **Fig. 3**) by eliminating the isolated pixels and conducted morphological operations, whose functionality has over lap with GAN.

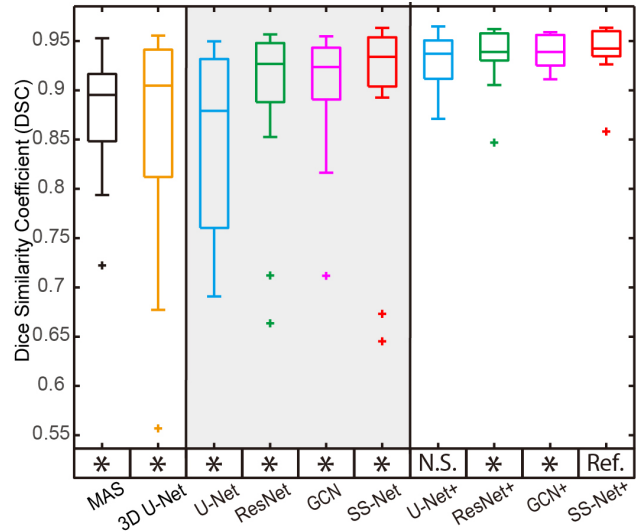

Fig. 10. The qualitative segmentation performance was presented as boxplots. The y axis indicated the DSC values, while the x axis reflected different methods. The statistical analyses were conducted between the proposed SS-Net+ method (marked as reference "Ref.") with other approaches. Statistically significant, differences were marked with a "*" symbol. Non-significant differences were indicated with "N.S.

The qualitative and quantitative results for the last epoch were presented in **Fig. 8** and **9**. In **Fig. 8**, segmentations on three subjects (highest, median, and lowest DSC in SS-Net+ methods) as well as the corresponding manual segmentation were showed. In **Fig. 9**, the boxplot of DSC for different

TABLE II
PERFORMANCE ON VOLUME AND SURFACE MEASUREMENTS

| | | MAS | 3D U-Net | U-Net | ResNet | GCN | SS-Net | U-Net+ | ResNet+ | GCN+ | SS-Net+ |
|---|---|---|---|---|---|---|---|---|---|---|---|
| DSC | median | 0.895 | 0.904 | 0.879 | 0.926 | 0.923 | 0.933 | 0.937 | 0.938 | 0.938 | **0.942** |
| | mean±std | 0.877±0.06 | 0.86±0.116 | 0.849±0.098 | 0.894±0.088 | 0.901±0.064 | 0.9±0.1 | 0.929±0.03 | 0.936±0.029 | 0.939±0.017 | **0.941**±0.026 |
| MSD | median | 4.13 | 4.453 | 7.427 | 4.318 | 5.411 | 3.043 | 2.528 | 1.667 | 1.712 | **1.484** |
| | mean±std | 5.137±3.154 | 5.916±5.378 | 8.435±3.51 | 7.114±6.973 | 8.721±8.529 | 4.871±3.547 | 2.82±1.435 | 2.474±1.762 | 2.304±1.319 | **2.294**±1.735 |
| HD | median | 44.327 | 38.104 | 239.354 | 146.009 | 160.191 | 84.192 | 39.48 | 22.81 | 22.5 | **22.345** |
| | mean±std | 50.2±28.2 | 71.2±60.3 | 246.7±38.0 | 144.3±55.3 | 156.9±52.1 | 103.3±59.5 | 34.4±11.7 | 28.8±14.2 | 28.4±12.6 | **27.2**±13.4 |

*DSC is Dice similarity coefficient, MSD is mean surface distance (mm), HD is Hausdorff distance (mm).

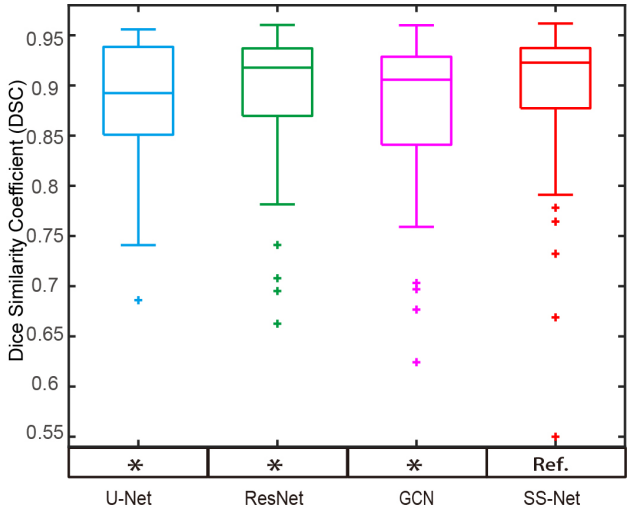

Fig. 11. The qualitative segmentation performance for 4-fold cross validation on all scans was presented as boxplots. The y axis indicated the DSC values, while the x axis reflected different methods. The quantitative results are provided in Table III. The statistical analyses were conducted between the proposed SS-Net method (marked as reference "Ref.") with other approaches. Statistically significant, differences were marked with a "*" symbol. Non-significant differences were indicated with "N.S".

TABLE III
PERFORMANCE ON CROSS VALIDATION

|  |  | U-Net | ResNet | GCN | SS-Net |
|---|---|---|---|---|---|
| DSC | median | 0.893 | 0.918 | 0.901 | **0.923** |
|  | mean±std | 0.875±0.102 | 0.894±0.067 | 0.878±0.076 | **0.897±0.074** |

*DSC is Dice similarity coefficient

methods were posed. In the left panel, the L-SIMPLE multi-atlas segmentation (MAS) and 3D U-Net were presented as baseline methods for multi-atlas and 3D DCNN network. The middle panel showed the DSC of U-Net, ResNet, GCN and SS-Net, while the right panel presented the DSC of U-Net+, ResNet+, GCN+, and SS-Net+. The SS-Net+ was used as reference method ("Ref."), which was compared with other methods using Wilcoxon signed rank test. The significant differences were marked as ("*"), while the remaining one was marked as ("N.S."). From the statistical analyses, the SS-Net+ achieved significantly better performance on DSC than other methods except U-Net+.

The volumetric and surface metrics for the last epoch were presented in **Table II**. The DSC, MSD, and HD were employed as measurements to evaluate different methods. Bold values indicated the best performance among different methods. From the results, the SS-Net+ achieved the best performance among all methods.

For a more complete analyses, the 4-fold cross-validation was also performed by splitting all 60 scans to 4 folds. Note that, the 45 training and 15 testing scans in the previous experiments were employed as fold 1. For each fold, 45 scans were used as training while the remaining 15 were used as validation. Each scan was used as a validation scan for only one time. The hyper-parameters for all folds were kept the same as the experiments in fold 1 with the same training epoch. The performance of U-Net, ResNet, GCN, and SS-Net on cross-validation results (60 validation scans from all folds) were provided in **Fig. 11** and **Table III**. The proposed method achieved superior segmentation performance. The differences between SS-Net ("Ref.") and other methods are significant from Wilcoxon signed rank test with p<0.05.

## VII. DISCUSSION

In this work, the Dice loss was employed as a canonical supervised learning loss rather than using weighted cross-entropy loss since the splenomegaly is not only an un-balanced per-voxel classification problem, but also has large variation for spleen volume size. Therefore, the segmentation performance was sensitive to the choice of weights. However, using Dice loss is able to avoid the subjective parameter choice and deal with large anatomical variation of splenomegaly.

Many components have been used in the proposed framework (e.g., large kernels, PatchGAN, 2D training etc.), The benefits of using large kernel has been presented in ResNet, GCN, and SS-Net in **Fig. 10** and **Table II**. ResNet used regular 3×3 kernels, GCN used dummy 7×7 kernels (1×7 and 7×1 kernels), and the proposed SS-Net used real 7×7 kernels. The results showed that the larger real kernels yield superior segmentation performance. The performance of using or without using PatchGAN was presented in **Fig. 10**. The PatchGAN helped the single-view SS-Net to achieve better performance for larger epochs. Using PatchGAN for multi-view SS-Net+ did not provide large improvement since the multi-view integration had taken care of some issues for GAN (e.g., isolated pixels and spatial corrections). The different performance between 2D and 3D training was shown in **Fig. 10** and **Table II**. The 2D training strategies yielded better performance compared with 3D training due to the limited number of training scans.

The ideas of using large kernels for skip-connection, GAN loss, and multi-view aggregation have been proposed from recent efforts. In this work, we proposed to use real 2D large kernels in skip connection rather than the previously used pseudo 2D large kernels. We used the batch-wise Dice loss rather than canonical slice-wise Dice loss, while using the unbalanced training loss to alleviate the saturation problem when incorporating the discriminator loss. We employed the multi-view aggregation to further improve the splenomegaly segmentation performance by combining union, open, and close morphological operations. Beyond improving each methodology component, this work integrates different components into a single pipeline to address the splenomegaly segmentation. From the application perspective, most of previous efforts on spleen segmentation have been made on CT scans, while limited efforts have been conducted on MRI spleen segmentation. To the best of our knowledge, this was first work that perform DCNN based splenomegaly segmentation using both T1w and T2w modalities simultaneously. Last, both 2D and 3D segmentation performance as well as multi-view integrations have been evaluated in this paper, which provides baseline performance for future efforts on MRI splenomegaly segmentation.

The limitation of this work is the small number of available splenomegaly scans. As a result, the 3D network yielded inferior performance compared with 2D networks. However, when acquiring more splenomegaly MRI scans (e.g., > 500), the performance of 3D networks might be leveraged. In SS-Net+, the three orthogonal views have been used to train three different models and three segmentations were integrated to the final segmentation. The multi-view images can be also used in training a 2.5D network directly, which has been successfully applied on different medical image segmentation applications, especially for the patch-based methods. The next meaning extension for this work would be design 2.5D based network structure in an end-to-end training manner.

In this study, the results of using balanced number of T1w and T2w in both training and testing have been shown in **Fig. 7**, **Fig.9** and **Table II**. The rationale of such design is due to the low generalizability of DCNN based method, which is also one major limitation for DCNN. Since we would like to apply the trained model to both clinical acquired T1w and T2w testing images, we keep both T1w and T2w used during training. As a result, we are able to segment a testing scan with T1w or T2w modality as we directly trained model using mixed modalities data, In the future, we would be able to achieve better performance by performing inter-modal intensity harmonization or image-to-image synthesis.

Another limitation of this work is the limited number of available training and testing MRI scans. In this study, 60 scans were split to 45 training and 15 testing as an internal validation, which also balanced two modalities (24 T1w / 21 T2w for training, while 8 T1w / 7 T2w for testing). Due to the limited resources on labeled MRI splenomegaly scans, external validations on MRI splenomegaly segmentation were not conducted in this work. For a more complete analyses, 4-fold cross validations were performed. Please note that the multi-fold cross validation with DCNN methods requires rather long training times. For instance, the 4-fold cross validation on four single-view methods for 50 epochs took >30 days computational time on a single GPU. When performing more epochs with larger size of cohort or with more baseline methods, the computational time would be even larger.

As different imaging modalities (e.g., T1w, T2w, fluid-attenuated inversion recovery (FLAIR) etc.) can be used for different clinical purposes, it is appealing to train a single deep neural network that is able to be applied to different modalities. Therefore, multi-modal training and testing have been explored in this study to evaluate the performance of splenomegaly segmentation that can be co-learned from both T1w and T2w scans. In existing experiments, the proposed method was trained by both modalities without particular technical operations such as specifying which imaging modalities were used in a batch. The segmentation performance could possibly be further improved by adding an extra image channel to encode the modality type during training, e.g. integrating ideas from recent work of [60]. Meanwhile, only two modalities (T1w and T2w) were included in the experiment. However, there are more imaging modalities in clinical practice, even with larger inter-modality variations than the used cohort. Therefore, it would be useful to perform further co-learning from T1w MRI, T2w MRI, contrast MRI, non-contrast MRI, FLAIR, and even CT. If successful, a single trained model could be used to address spleen segmentation on a variety of imaging modalities.

VIII. CONCLUSION

In this work, we introduced the deep convolutional networks to the application of MRI splenomegaly segmentation. To address the particular challenges of large anatomical variations of splenomegaly and large intensity variations of multi-modal MRI, the SS-Net were proposed using large convolutional kernels and adversarial learning. To further leverage the segmentation performance, the 2D models were extended to 2D+ models. In empirical validation (**Fig. 10** and **Table II**), 2D, 2D+ and 3D DCNN networks as well as state-of-the-art multi-atlas segmentation method have been evaluated. From empirical validation, the DCNN methods achieved higher median and mean DSC than multi-atlas segmentation with less outliers (e.g., DSC < 0.8). 2D+ strategy using three networks (axial, sagittal, and coronal) achieved higher median DSC performance than using one 2D network (on axial view). The proposed SS-Net achieved superior performance than baseline DCNN methods for both 2D and 2D+ models.

REFERENCES

[1]   P. A. McCormick and K. M. Murphy, "Splenomegaly, hypersplenism and coagulation abnormalities in liver disease," *Baillieres Best Pract Res Clin Gastroenterol,* vol. 14, pp. 1009-31, Dec 2000.

[2]   A. W. Woodruff, "Mechanisms involved in anaemia associated with infection and splenomegaly in the tropics," *Trans R Soc Trop Med Hyg,* vol. 67, pp. 313-28, 1973.

[3]   B. Klein, M. Stein, A. Kuten, M. Steiner, D. Barshalom, E. Robinson*, et al.*, "Splenomegaly and solitary spleen metastasis in solid tumors," *Cancer,* vol. 60, pp. 100-2, Jul 01 1987.

[4]   R. G. Hosey, C. G. Mattacola, V. Kriss, T. Armsey, J. D. Quarles, and J. Jagger, "Ultrasound assessment of spleen size in collegiate athletes," *Br J Sports Med,* vol. 40, pp. 251-4; discussion 251-4, Mar 2006.

[5]   I. De Odorico, K. A. Spaulding, D. H. Pretorius, A. S. Lev-Toaff, T. B. Bailey, and T. R. Nelson, "Normal splenic volumes estimated using three-dimensional ultrasonography," *Journal of ultrasound in medicine,* vol. 18, pp. 231-236, 1999.

[6]   P. Prassopoulos, M. Daskalogiannaki, M. Raissaki, A. Hatjidakis, and N. Gourtsoyiannis, "Determination of normal splenic volume on computed tomography in relation to age, gender and body habitus," *Eur Radiol,* vol. 7, pp. 246-8, 1997.

[7]   A. S. Bezerra, G. D'Ippolito, S. Faintuch, J. Szejnfeld, and M. Ahmed, "Determination of splenomegaly by CT: is there a place for a single measurement?," *AJR Am J Roentgenol,* vol. 184, pp. 1510-3, May 2005.

[8]   M. G. Linguraru, J. K. Sandberg, E. C. Jones, and R. M. Summers, "Assessing splenomegaly: automated volumetric analysis of the spleen," *Acad Radiol,* vol. 20, pp. 675-84, Jun 2013.

[9] A. S. Bezerra, G. D'Ippolito, S. Faintuch, J. Szejnfeld, and M. Ahmed, "Determination of splenomegaly by CT: is there a place for a single measurement?," *American Journal of Roentgenology,* vol. 184, pp. 1510-1513, 2005.

[10] C. Thomsen, P. Josephsen, H. Karle, E. Juhl, P. G. Sorensen, and O. Henriksen, "Determination of T1- and T2-relaxation times in the spleen of patients with splenomegaly," *Magn Reson Imaging,* vol. 8, pp. 39-42, 1990.

[11] M. Mazonakis, J. Damilakis, T. Maris, P. Prassopoulos, and N. Gourtsoyiannis, "Estimation of spleen volume using MR imaging and a random marking technique," *European Radiology,* vol. 10, pp. 1899-1903, 2000.

[12] A. Mihaylova and V. Georgieva, "A Brief Survey of Spleen Segmentation in MRI and CT Images," *International Journal,* vol. 5, 2016.

[13] P. Campadelli, S. Pratissoli, E. Casiraghi, and G. Lombardi, "Automatic abdominal organ segmentation from CT images," *ELCVIA: electronic letters on computer vision and image analysis,* vol. 8, pp. 001-14, 2009.

[14] P. Campadelli, E. Casiraghi, and S. Pratissoli, "A segmentation framework for abdominal organs from CT scans," *Artif Intell Med,* vol. 50, pp. 3-11, Sep 2010.

[15] X. Chen, J. K. Udupa, U. Bagci, Y. Zhuge, and J. Yao, "Medical image segmentation by combining graph cuts and oriented active appearance models," *IEEE Trans Image Process,* vol. 21, pp. 2035-46, Apr 2012.

[16] A. Behrad and H. Masoumi, "Automatic spleen segmentation in MRI images using a combined neural network and recursive watershed transform," in *Neural Network Applications in Electrical Engineering (NEUREL), 2010 10th Symposium on*, 2010, pp. 63-67.

[17] M. G. Linguraru, J. K. Sandberg, Z. Li, F. Shah, and R. M. Summers, "Automated segmentation and quantification of liver and spleen from CT images using normalized probabilistic atlases and enhancement estimation," *Med Phys,* vol. 37, pp. 771-783, 2010.

[18] Z. Xu, R. P. Burke, C. P. Lee, R. B. Baucom, B. K. Poulose, R. G. Abramson*, et al.*, "Efficient multi-atlas abdominal segmentation on clinically acquired CT with SIMPLE context learning," *Medical image analysis,* vol. 24, pp. 18-27, 2015.

[19] J. Liu, Y. Huo, Z. Xu, A. Assad, R. G. Abramson, and B. A. Landman, "Multi-atlas spleen segmentation on CT using adaptive context learning," in *Medical Imaging 2017: Image Processing*, 2017, p. 1013309.

[20] N. Lay, N. Birkbeck, J. Zhang, and S. K. Zhou, "Rapid multi-organ segmentation using context integration and discriminative models," in *International Conference on Information Processing in Medical Imaging*, 2013, pp. 450-462.

[21] M. P. Heinrich and M. Blendowski, "Multi-organ segmentation using vantage point forests and binary context features," in *International Conference on Medical Image Computing and Computer-Assisted Intervention*, 2016, pp. 598-606.

[22] E. C. Lin, "Radiation risk from medical imaging," in *Mayo Clinic Proceedings*, 2010, pp. 1142-1146.

[23] S. W. Farraher, H. Jara, K. J. Chang, A. Hou, and J. A. Soto, "Liver and Spleen Volumetry with Quantitative MR Imaging and Dual-Space Clustering Segmentation 1," *Radiology,* vol. 237, pp. 322-328, 2005.

[24] J. Wu, *An automated human organ segmentation technique for abdominal magnetic resonance images*: McMaster University, 2010.

[25] O. Pauly, B. Glocker, A. Criminisi, D. Mateus, A. M. Möller, S. Nekolla*, et al.*, "Fast multiple organ detection and localization in whole-body MR Dixon sequences," in *International Conference on Medical Image Computing and Computer-Assisted Intervention*, 2011, pp. 239-247.

[26] Y. Huo, J. Liu, Z. Xu, R. L. Harrigan, A. Assad, R. G. Abramson*, et al.*, "Multi-atlas segmentation enables robust multi-contrast MRI spleen segmentation for splenomegaly," in *Medical Imaging 2017: Image Processing*, 2017, p. 101330A.

[27] Y. Huo, J. Liu, Z. Xu, R. L. Harrigan, A. Assad, R. G. Abramson*, et al.*, "Robust Multicontrast MRI Spleen Segmentation for Splenomegaly Using Multi-Atlas Segmentation," *IEEE Transactions on Biomedical Engineering,* vol. 65, pp. 336-343, 2018.

[28] J. Long, E. Shelhamer, and T. Darrell, "Fully convolutional networks for semantic segmentation," in *Proceedings of the IEEE conference on computer vision and pattern recognition*, 2015, pp. 3431-3440.

[29] O. Ronneberger, P. Fischer, and T. Brox, "U-net: Convolutional networks for biomedical image segmentation," in *International Conference on Medical image computing and computer-assisted intervention*, 2015, pp. 234-241.

[30] V. Badrinarayanan, A. Kendall, and R. Cipolla, "Segnet: A deep convolutional encoder-decoder architecture for image segmentation," *IEEE transactions on pattern analysis and machine intelligence,* vol. 39, pp. 2481-2495, 2017.

[31] L.-C. Chen, G. Papandreou, I. Kokkinos, K. Murphy, and A. L. Yuille, "Deeplab: Semantic image segmentation with deep convolutional nets, atrous convolution, and fully connected crfs," *IEEE transactions on pattern analysis and machine intelligence,* vol. 40, pp. 834-848, 2018.

[32] H. Noh, S. Hong, and B. Han, "Learning deconvolution network for semantic segmentation," in *Proceedings of the IEEE International Conference on Computer Vision*, 2015, pp. 1520-1528.

[33] Z. Xu, Y. Huo, J. Park, B. Landman, A. Milkowski, S. Grbic*, et al.*, "Less is More: Simultaneous View Classification and Landmark Detection for Abdominal Ultrasound Images," in *International Conference on Medical Image Computing and Computer-Assisted Intervention*, 2018, pp. 711-719.

[34] H. R. Roth, A. Farag, L. Lu, E. B. Turkbey, and R. M. Summers, "Deep convolutional networks for pancreas segmentation in CT imaging," *arXiv preprint arXiv:1504.03967,* 2015.
[35] P. Hu, F. Wu, J. Peng, P. Liang, and D. Kong, "Automatic 3D liver segmentation based on deep learning and globally optimized surface evolution," *Physics in medicine and biology,* vol. 61, p. 8676, 2016.
[36] F. Lu, F. Wu, P. Hu, Z. Peng, and D. Kong, "Automatic 3D liver location and segmentation via convolutional neural network and graph cut," *International journal of computer assisted radiology and surgery,* vol. 12, pp. 171-182, 2017.
[37] P. Hu, F. Wu, J. Peng, Y. Bao, F. Chen, and D. Kong, "Automatic abdominal multi-organ segmentation using deep convolutional neural network and time-implicit level sets," *International journal of computer assisted radiology and surgery,* vol. 12, pp. 399-411, 2017.
[38] Y. Huo, Z. Xu, S. Bao, C. Bermudez, A. J. Plassard, J. Liu*, et al.*, "Splenomegaly Segmentation using Global Convolutional Kernels and Conditional Generative Adversarial Networks," *arXiv preprint arXiv:1712.00542,* 2017.
[39] C. Peng, X. Zhang, G. Yu, G. Luo, and J. Sun, "Large Kernel Matters--Improve Semantic Segmentation by Global Convolutional Network," *arXiv preprint arXiv:1703.02719,* 2017.
[40] Y. Zhou, L. Xie, W. Shen, Y. Wang, E. K. Fishman, and A. L. Yuille, "A fixed-point model for pancreas segmentation in abdominal CT scans," in *International Conference on Medical Image Computing and Computer-Assisted Intervention*, 2017, pp. 693-701.
[41] Y. Zhou, L. Xie, E. K. Fishman, and A. L. Yuille, "Deep supervision for pancreatic cyst segmentation in abdominal CT scans," in *International Conference on Medical Image Computing and Computer-Assisted Intervention*, 2017, pp. 222-230.
[42] J. Cai, L. Lu, Y. Xie, F. Xing, and L. Yang, "Improving deep pancreas segmentation in CT and MRI images via recurrent neural contextual learning and direct loss function," *arXiv preprint arXiv:1707.04912,* 2017.
[43] B. Zhou, A. Khosla, A. Lapedriza, A. Oliva, and A. Torralba, "Object detectors emerge in deep scene cnns," *arXiv preprint arXiv:1412.6856,* 2014.
[44] I. Goodfellow, J. Pouget-Abadie, M. Mirza, B. Xu, D. Warde-Farley, S. Ozair*, et al.*, "Generative adversarial nets," in *Advances in neural information processing systems*, 2014, pp. 2672-2680.
[45] E. L. Denton, S. Chintala, and R. Fergus, "Deep generative image models using a laplacian pyramid of adversarial networks," in *Advances in neural information processing systems*, 2015, pp. 1486-1494.
[46] C. Ledig, L. Theis, F. Huszár, J. Caballero, A. Cunningham, A. Acosta*, et al.*, "Photo-realistic single image super-resolution using a generative adversarial network," *arXiv preprint,* 2016.
[47] P. Isola, J.-Y. Zhu, T. Zhou, and A. A. Efros, "Image-to-image translation with conditional adversarial networks," *arXiv preprint,* 2017.
[48] P. Luc, C. Couprie, S. Chintala, and J. Verbeek, "Semantic segmentation using adversarial networks," *arXiv preprint arXiv:1611.08408,* 2016.
[49] J. Lafferty, A. McCallum, and F. C. Pereira, "Conditional random fields: Probabilistic models for segmenting and labeling sequence data," 2001.
[50] J. A. Sethian, "A fast marching level set method for monotonically advancing fronts," *Proceedings of the National Academy of Sciences,* vol. 93, pp. 1591-1595, 1996.
[51] B. Van Ginneken, A. F. Frangi, J. J. Staal, B. M. ter Haar Romeny, and M. A. Viergever, "Active shape model segmentation with optimal features," *IEEE transactions on medical imaging,* vol. 21, pp. 924-933, 2002.
[52] K. He, X. Zhang, S. Ren, and J. Sun, "Deep residual learning for image recognition," in *Proceedings of the IEEE conference on computer vision and pattern recognition*, 2016, pp. 770-778.
[53] Ö. Çiçek, A. Abdulkadir, S. S. Lienkamp, T. Brox, and O. Ronneberger, "3D U-Net: learning dense volumetric segmentation from sparse annotation," in *International Conference on Medical Image Computing and Computer-Assisted Intervention*, 2016, pp. 424-432.
[54] H. R. Roth, L. Lu, N. Lay, A. P. Harrison, A. Farag, A. Sohn*, et al.*, "Spatial aggregation of holistically-nested convolutional neural networks for automated pancreas localization and segmentation," *Medical image analysis,* vol. 45, pp. 94-107, 2018.
[55] D. P. Kingma and J. Ba, "Adam: A method for stochastic optimization," *arXiv preprint arXiv:1412.6980,* 2014.
[56] M. P. Heinrich, M. Jenkinson, M. Brady, and J. A. Schnabel, "MRF-based deformable registration and ventilation estimation of lung CT," *IEEE Trans Med Imaging,* vol. 32, pp. 1239-48, Jul 2013.
[57] H. Z. Wang, J. W. Suh, S. R. Das, J. B. Pluta, C. Craige, and P. A. Yushkevich, "Multi-Atlas Segmentation with Joint Label Fusion," *Ieee Transactions on Pattern Analysis and Machine Intelligence,* vol. 35, pp. 611-623, Mar 2013.
[58] Z. Xu, R. P. Burke, C. P. Lee, R. B. Baucom, B. K. Poulose, R. G. Abramson*, et al.*, "Efficient multi-atlas abdominal segmentation on clinically acquired CT with SIMPLE context learning," *Med Image Anal,* vol. 24, pp. 18-27, Aug 2015.
[59] F. Wilcoxon, "Individual comparisons by ranking methods," *Biometrics bulletin,* pp. 80-83, 1945.
[60] Y. Choi, M. Choi, M. Kim, J.-W. Ha, S. Kim, and J. Choo, "Stargan: Unified generative adversarial networks for multi-domain image-to-image translation," *arXiv preprint,* vol. 1711, 2017.